\documentclass[10pt,twocolumn,letterpaper]{article}

\usepackage[pagenumbers]{cvpr}

\usepackage{graphicx}
\usepackage{amsmath}
\usepackage{amssymb}
\usepackage{booktabs}
\usepackage{fmsymbols}
\usepackage[ruled,vlined,linesnumbered,resetcount]{algorithm2e}

\usepackage[pagebackref,breaklinks,colorlinks,bookmarks=false]{hyperref}

\usepackage[capitalize]{cleveref}
\crefname{section}{Sec.}{Secs.}
\Crefname{section}{Section}{Sections}
\Crefname{table}{Table}{Tables}
\crefname{table}{Tab.}{Tabs.}

\makeatletter
\renewcommand\paragraph{\@startsection{paragraph}
                                        {4}
                                        {\z@}%
                                        {0.5em}
                                        {-1em}%
                                        {\normalfont\normalsize\bfseries}}
\makeatother

\begin{document}

\title{Progressive Minimal Path Method with Embedded CNN}

\author{Wei Liao\\
Independent Researcher\\
{\tt\small liaowei.post@gmail.com}
}
\maketitle

\begin{abstract}

We propose Path-CNN, a method for the segmentation of centerlines of tubular structures by embedding convolutional neural networks (CNNs) into the progressive minimal path method.
Minimal path methods are widely used for topology-aware centerline segmentation, but usually these methods rely on weak, hand-tuned image features.
In contrast, CNNs use strong image features which are learned automatically from images.
But CNNs usually do not take the topology of the results into account, and often require a large amount of annotations for training.
We integrate CNNs into the minimal path method, so that both techniques benefit from each other:
CNNs employ learned image features to improve the determination of minimal paths, while the minimal path method ensures the correct topology of the segmented centerlines, provides strong geometric priors to increase the performance of CNNs, and reduces the amount of annotations for the training of CNNs significantly.
Our method has lower hardware requirements than many recent methods.
Qualitative and quantitative comparison with other methods shows that Path-CNN achieves better performance, especially when dealing with tubular structures with complex shapes in challenging environments.
\end{abstract}

\section{Introduction}
Topology-aware centerline segmentation for tubular structures plays a crucial role in computer vision.
One of its most important application areas is the quantitative analysis of roads and rivers in satellite images for measurement, planning, or navigation.
These are challenging tasks due to the complex shape of roads and rivers, and the high variability of their environment.

When using common methods for object segmentation for this task, usually a binary mask of the tubular structure is computed in the first step.
After that, post-processing, often based on heuristics, is necessary, in order to determine the centerline and to deal with small gaps on the tubular structure due to noise or image clutter.
In contrast, \emph{minimal path methods} based on Dijkstra's algorithms \cite{Dijkstra59:NM} or the fast marching method \cite{Cohen_IJCV_1997} provide a more elegant solution.
As in most minimal path methods, we assume that the start point $\fmstartpoint$ and end point $\fmendpoint$ of the centerline are given, and focus on the determination of the path itself.
Often, the start and end points are automatically obtained using application specific methods, such as \cite{Mosinska_PAMI_2020}.
Minimal path methods allow finding the best path as the global optimum of a cost function, while inherently enforcing strict line-topology, i.e., the result is always a sequence of coordinates of points on the centerline.
Also, small gaps on the path can be completed automatically.
However, minimal path methods usually utilize hand-tuned image features, such as differential measures \cite{Frangi1998, Law_ECCV_2008}, to distinguish between tubular structures and the background.
Such features are efficient to compute, but they are relatively weak and may lead to \emph{short cuts} for images containing challenging environments.
\begin{figure}[!t]
  \hspace*{-0.3cm}
  \centering
  \setlength{\tabcolsep}{3pt}
  \begin{tabular}{cc}
    \includegraphics[width=3.4cm]{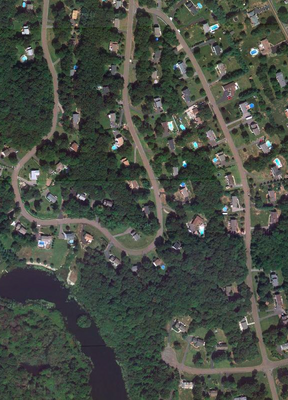}&
    \includegraphics[width=3.4cm]{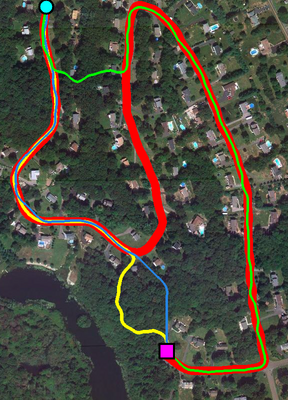}\\
    (a) Input image & (b) Segmented centerlines
  \end{tabular}
  \vspace*{-9pt}
  \caption{Short cuts due to complex centerline geometry.
    (a) The input is a satellite image of roads.
    (b) Start point (magenta box) and end point (cyan circle) of a road are given.
    Results of a previous method using only hand-tuned tubularity measure (yellow line) and two methods using CNNs (blue and green lines) contain short cuts.
    Our approach achieves the correct centerline (red line), although it is much longer than the results with short cuts.
  }
  \label{pic:epfl_shortcut}
\end{figure}
With convolutional neural networks (CNNs), stronger image features can be extracted automatically from images.
Although these features alone cannot ensure the topology of the centerlines, they can be used to better classifying the pixels and thus improve the results of minimal path methods.
However, CNNs often require large amounts of annotated training data, which can be expensive to obtain.
If only limited annotations are available, often CNN-based methods cannot be fully trained, and thus they also result in short cuts.
Examples of short cuts are shown in \cref{pic:epfl_shortcut}.

In this work, we propose a novel method, Path-CNN, to embed CNNs into the progressive minimal path method \cite{Liao_PAMI_2018_Final}, so that these two techniques operate \emph{alternately} and benefit from each other:
On the one hand, we use CNNs to learn image features automatically so that a wider variety of short cuts can be detected, and therefore minimal paths can be better determined.
Instead of learning features for \emph{isolated pixels} as in most previous approaches, our CNNs learn features for \emph{rectified patches} along \emph{paths}.
On the other hand, the progressive minimal path method not only ensures the line-topology of the results, but also provides strong geometric priors, which are used in turn to reduce the number of training samples for CNNs significantly.
Although we only use centerlines as training data, our method not only determines centerlines, but also produces binary segmentation masks for tubular structures.
To the best of our knowledge, this is the first approach to employ such geometric priors for CNNs to segment centerlines of tubular structures.
Compared with most other approaches based on deep learning, our method has lower requirements not only for the amount of annotations and but also for hardware.

\section{Related Work}
In this section, we review the two main components of minimal path methods: Image features and minimal path computation.
We also emphasize the differences between our method and other recent methods for road extraction.

\paragraph{Image features}
Minimal path methods often use tubularity measures as image features.
Such measures can be interpreted as the probability that a pixel belongs to a tubular structure.
Widely used features, such as Hessian-based measures \cite{Frangi1998,Sato1998} or flux-based features \cite{Law_ECCV_2008,Tueretken_ICCV_2013}, are hand-tuned.
There are also learning-based features.
For example, \cite{Sironi_PAMI_2015b} uses features based on decision trees, and \cite{Mnih_ECCV_2010} uses a CNN.
However, these features are learned for isolated \emph{pixels}.
Also, these features are \emph{static}, i.e., they are computed \emph{before} the minimal path computation starts, and remain constant thereafter.
In contrast, we use \emph{path-based features}.
While being stronger than their pixel-based counterparts, such features can only be computed \emph{dynamically}, i.e., \emph{during} the minimal path computation.

\paragraph{Minimal path computation}
There are different methods to overcome short cuts when computing minimal paths.
With domain-lifting, additional dimensions are introduced into the parameter space to represent more features such as line width or orientation \cite{Pechaud2009, Li2007}, but the computational cost increases significantly with the dimensionality.
In \cite{Benmansour_IJCV_2011,Chen_TIP_2019}, fast marching methods with anisotropic features are proposed, but such features require more complex numerical schemes than commonly used isotropic features.
We use Dijkstra's algorithm, in which both anisotropic and isotropic features are used in the same way, without the need for further numerical schemes.
In \cite{Ulen_PAMI_2015}, more complex graph structures are used to represent higher-order constraints such as curvature or torsion, but this results in high computation time and limited feature types.
In \cite{Benmansour2009b,Kaul-Yezzi-Tsai_PAMI_2012}, additional keypoints are inserted heuristically, and \emph{progressive minimal path methods} \cite{Liao_PAMI_2018_Final,Chen_TIP_2019} employ path-based features computed on-the-fly.
But these approaches still rely on hand-tuned features based on appearance or geometry.
Also, the path-based features in these approaches are still derived from pixel-based features.
In contrast, we use dynamic features which are learned directly using paths.
Furthermore, our method can handle a wider variety of short cuts in a uniform way.

\paragraph{Methods for road extraction}

For road extraction using satellite images, there exist recent approaches (e.g., \cite{Mattyus_ICCV_2017,Bastani_CVPR_2018,Mosinska_PAMI_2020}).
However, there are important distinctions between these methods and ours.
First, existing methods usually require large amounts of training data.
For example, \cite{Mosinska_PAMI_2020} uses one of the largest public datasets of road images \cite{Bastani_CVPR_2018}, and employs \abbrunet to obtain features for further refinement.
In contrast, by combining minimal path methods and CNNs, our method requires much less annotated training data than \abbrunet and most other architectures, i.e., our method performs better if the available annotation is very sparse.
Second, many models, such as \cite{Bastani_CVPR_2018}, have relatively high requirements for hardware, whereas our model can be trained and deployed efficiently using only 2GB GPU memory.
Third, previous methods were only applied to extract roads in urban or suburban areas, while our method has been used also for roads in other environments and rivers.
Furthermore, although our method is trained only with centerline annotations, it also produces a classification of each pixel in the image, which corresponds to a binary segmentation.
In this way, further properties of the tubular structures, such as width or area, can also be determined.

\section{Minimal Path Framework}
Segmentation of tubular structures can be formulated naturally using the minimal path framework, which relies on Dijkstra's algorithm in discrete cases, or the fast marching method in continuous cases.
Our approach focuses on Dijkstra's algorithm.
Let image \image induce a graph $\dijGraph = (\dijVertex, \dijEdge)$, where \dijVertex and \dijEdge are the sets of \emph{vertices} and \emph{edges}, respectively.
Each vertex corresponds to a pixel in $\image$, and vertices of neighboring pixels are connected by edges.
Function $\dijW$ uses image features to assign positive \emph{weights} to the edges.
A \emph{path} \mypath is a sequence of vertices $\{v_0, v_1, \dots, v_{|\mypath|}\}$.
Given a start point \fmstartpoint and an end point \fmendpoint, the minimal path \minimalpath, which corresponds to the centerline of a tubular structure in the image, can be determined by minimizing the following cost function
\begin{equation}
\minimalpath = \argmin_{\mypath\in\allpaths(\fmstartpoint,\fmendpoint)}\sum_{i=1}^{|\mypath|} \dijW[e_{v_{i-1}, v_i}],
\label{eqn:cost_minpath}
\end{equation}
where $\allpaths(\fmstartpoint,\fmendpoint)$ is the set of all paths connecting \fmstartpoint and \fmendpoint, and $v_i$ is the $i$-th vertex on the path \mypath.
Consecutive vertices $v_{i-1}$ and $v_i$ on \mypath are connected by edge $e_{v_{i-1}, v_i}$.

To minimize \eqref{eqn:cost_minpath}, we use a unified formulation of Dijkstra's algorithm adapted from \cite{Cormen2009}, as shown in \cref{alg:dijkstra}.
For each vertex $u$, $\dijPrev(u)$ specifies its \emph{predecessor} in the path, while $\dijDist(u)$ is the \emph{path weight}, i.e., sum of weights of all edges on the path between $u$ and the start point \fmstartpoint.
$\dijN(u)$ denotes the set of neighboring vertices of $u$.
We introduce a new function \dijAdaptNeighborEdge to transform \dijWinit.
In the standard Dijkstra's algorithm, \dijAdaptNeighborEdge just returns the initial weight $\dijWinit[u, v]$ for edge $e_{u,v}$, i.e., $\dijW = \dijWinit$.
An improved version of \dijAdaptNeighborEdge is proposed in \cref{sec:pathcnn_method} below.
\dijVertex is divided into two disjoint sets \dijQ and \dijS.
For each vertex $u$ in \dijS, $\dijDist(u)$ and $\dijPrev(u)$ are finalized, while in \dijQ, they may still be updated.
In each iteration of the main loop, the vertex $u$ with minimum path weight is moved from \dijQ to \dijS, and it is checked for each neighbor $v\in\dijN(u)$ whether $d(v)$ can be reduced by reaching $v$ via $e_{u,v}$.
If this is the case, then $u$ becomes predecessor of $v$.
Once the end point \fmendpoint is added to \dijS, \minimalpath can be extracted by starting at \fmendpoint, and recursively looking up the predecessors using \dijPrev, until the start point \fmstartpoint is reached.
The centerlines obtained in this way automatically have strict \emph{line-topology}.
\begin{algorithm}[!t]
  \LinesNumbered
  \SetKwData{result}{result}
  \SetKw{init}{Initialization}
  \KwIn{Start point $\dijStart$, end point $\dijEnd$, image $\image$, graph $\dijGraph = (\dijVertex, \dijEdge)$, initial edge weights $\dijWinit$,}
  \KwOut{predecessor function $\dijPrev$}
  \SetKwFunction{GetInitialWeight}{\dijGetInitialWeight}
  \SetKwFunction{SetupGraph}{\dijSetupGraph}
  \SetKwFunction{AdaptNeighborEdge}{\dijAdaptNeighborEdge}
  \For{each $\position\in\dijVertex$}{
    $\dijPrev[\position] \assign \dijNone$; $\dijDist[\position] \assign \infty$\;
  }
  $\dijDist[\dijStart] \assign 0$; $\dijQ \assign \dijVertex$; $\dijS \assign \emptyset$\;

  \While(\tcp*[h]{Main loop}){$\fmendpoint \notin \dijS$}{\label{func:main_loop}
    $u \assign \argmin_{\position\in\dijQ}\dijDist(\position)$\;
    $\dijQ \assign \dijQ - \{u\}$; $\dijS \assign \dijS \cup \{u\}$\;
    \For{each $v \in \dijN(u)$}{
      $\dijW[e_{u,v}] = \AdaptNeighborEdge(u, v, \dijWinit, \dijPrev, \image)$\;
      \If{$\dijDist[v] > \dijDist[u] + \dijW[e_{u,v}]$}{
        $\dijDist[v] \assign \dijDist[u] + \dijW[e_{u,v}]$\;
        $\dijPrev[v] \assign u$\;
      }
    }
  }
  \Return $\dijPrev$
  \caption{Unified Dijkstra's algorithm.
  }
  \label{alg:dijkstra}
\end{algorithm}

Usually, the initial weight \dijWinit for edge $e$ is defined as
\begin{equation}
\dijWinit[e] = \frac{1}{\vesselmap(e) + \epsilon} + \lambda\cdot \mylength(e),
\end{equation}
where $\vesselmap(e)$ is a tubularity measure, and \smallconst is a small constant to avoid division by zero.
The term $\lambda\cdot\mylength(e)$ controls the smoothness of the path, where $\mylength(e)$ is the Euclidean length of $e$, and \smoothweight is a constant.
In our method, the tubularity measure \cite{Frangi1998} is used.

\section{Path-CNN Method}
\label{sec:pathcnn_method}
In this section, we propose Path-CNN, a new approach to solving a common problem of minimal path methods: The short cut problem.
To do so, we embed CNNs into the progressive minimal path method \cite{Liao_PAMI_2018_Final} in such a way that these two techniques naturally complement each other to achieve better performance.

\subsection{Taxonomy of Short Cuts}
\begin{figure}[!t]
  \centering
  \begin{tabular}{ccc}
    \includegraphics[angle=90,origin=c,width=2.2cm]{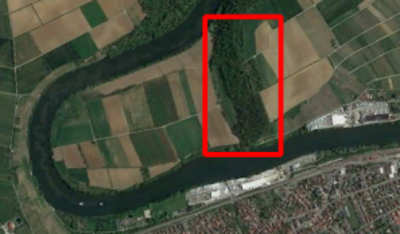}&
    \includegraphics[angle=90,origin=c,width=2.2cm]{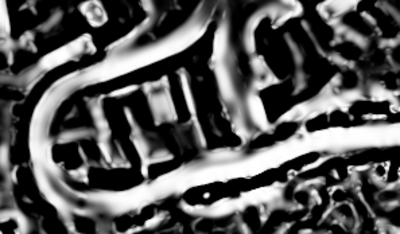}&
    \includegraphics[angle=90,origin=c,width=2.2cm]{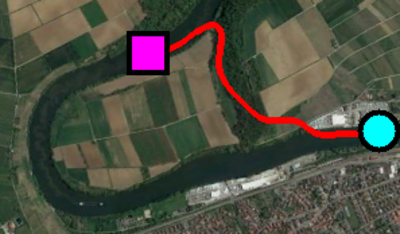}\\
    (a) Input image & (b) Tubularity & (c) Short cut
  \end{tabular}
\vspace*{-5pt}
  \caption{A short cut due to similar appearance of foreground and background.
    The region in the red box in (a) has similar tubularity as the foreground in (b), leading to a short cut in (c).
  }
  \label{pic:river_shortcut}
\end{figure}

Short cuts are incorrect centerlines found by minimal path methods.
There are mainly two reasons for short cuts, which we refer to as Type 1 and Type 2, respectively.
In cases of Type 1, the correct path may be very long and curved, so that a wrong but shorter connection, despite running through the background, still achieves a lower cost \eqref{eqn:cost_minpath}, such as the examples in \cref{pic:epfl_shortcut}b.
In cases of Type 2, the background may appear very similar to the foreground.
For example, in \cref{pic:river_shortcut}a, the red box indicates a background region similar to the river nearby.
This region has also high tubularity (\cref{pic:river_shortcut}b), leading to a short cut (\cref{pic:river_shortcut}c).
Most previous methods only attempt to avoid short cuts of Type 1, and some methods deal with Type 2 under certain assumptions for the geometry or appearance of the tubular structures (e.g., \cite{Chen_TIP_2019}).
In contrast, our method takes both types into account in a general and uniform way.

\subsection{Path Classification}
\label{sec:path_classification}
To avoid short cuts, we add a step into Dijkstra's algorithm: A CNN is applied to classify image patches extracted using local paths, and the classification result is used to detect possible short cuts.

\paragraph{Local paths}
Following \cite{Liao_PAMI_2018_Final}, the \emph{local path} $\dijPath(u)$ at a vertex $u\in\dijS$ is defined as the path of a \emph{constant} length \localpathlength.
$\dijPath(u)$ can be determined by starting at $u$ with an empty path, and recursively looking up the predecessors using \dijPrev, until the path length reaches \localpathlength.
Intuitively, we can avoid \minimalpath with short cuts by detecting and avoiding its segment \dijPath which contains such short cuts.
Using our method described below, we are able to use local paths to cope with short cuts of Types 1 and 2 by applying a single CNN.

\paragraph{CNN and path-based features}
To use path-based features, we introduce three new operations into the function \dijAdaptNeighborEdge: \emph{Cropping} of tubular patches along local paths, \emph{rectification} of these patches, and \emph{classification} of rectified patches using a CNN.
All these steps must be computed on-the-fly, i.e., \emph{during} the minimal path computation, since local paths need to be computed using \dijPrev in the set \dijS, but \dijS is non-empty only after the main loop of \cref{alg:dijkstra} has started in \cref{func:main_loop}.

\begin{algorithm}[!t]
  \LinesNumbered
  \SetKwData{result}{result}
  \SetKw{init}{Initialization}
  \KwIn{vertices $u$ and $v$, weight $\dijWinit$, $\dijPrev$, image $\image$}
  \KwOut{adapted weight $\dijW_a$}
  \SetKwFunction{GetLocalPath}{\dijGetLocalPath}
  \SetKwFunction{Crop}{\dijCrop}
  \SetKwFunction{Rectify}{\dijRectify}
  \SetKwFunction{Classify}{\dijClassify}

  $\dijPath \assign \GetLocalPath(u, \dijPrev)$\;
  $ \dijP \assign \Crop(\dijPath, \image)$\;
  $\dijRect \assign \Rectify(\dijPath, \dijP)$\;
  $\dijClass \assign \Classify(\dijRect)$\label{func:classify}\;

  \eIf{$\dijClass = \dijFG$}{
    $\dijW_a = \dijWinit[e_{u,v}]$\;
  }(\tcp*[h]{$\dijClass = \dijBG$, add penalty}){
    $\dijW_a = \dijWinit[e_{u,v}] + \dijPenalty$\;
  }

  \Return $\dijW_a$
  \caption{\dijAdaptNeighborEdge for Path-CNN.}
  \label{alg:pathcnn}
\end{algorithm}

Details of \dijAdaptNeighborEdge are shown in \cref{alg:pathcnn}.
The initial edge weights \dijWinit are computed using tubularity measure.
Depending on the classification result of the CNN, the final weight \dijW is either equal to \dijWinit, or much higher than \dijWinit.
First, we extract the \emph{local path} $\dijPath$ at vertex $u$.
Then, the image region along \dijPath is \emph{cropped} to a tubular image patch \dijP with constant width, so that \dijPath is the centerline of \dijP.
In the subsequent step of \emph{rectification}, the tubular patch \dijP is transformed into a rectangular patch by warping it along its centerline, and rotated into a canonical orientation, resulting in a rectified patch \dijRect.
Then \dijRect is \emph{classified} by the CNN, which has been trained using rectified image patches in the canonical orientation, instead of using non-rectified image patches of arbitrary orientations.
Subsequent steps depend on the classification result of the CNN:
If \dijRect is classified as foreground, we conclude that $u$ (the start point of \dijPath) is inside certain tubular structures.
In this case, the weight of the edge between $u$ and $v$ does not change.
This is the same as in the standard Dijkstra's algorithm.
On the other hand, if \dijRect is classified as background, then we conclude that $u$ is not inside a tubular structure, and consequently $\dijPath$ is more likely to be part of a short cut than part of the final minimal path \minimalpath, i.e., a possible short cut is detected at $u$.
In this case, we increase the weight of the edge $e_{u,v}$ from its initial value $\dijWinit$ by a penalty \dijPenalty, which is a large positive number.
So, even if $v$ eventually turns out to be on \minimalpath, the probability that $u$ is predecessor of $v$ on \minimalpath is significantly reduced due to \dijPenalty, since $v$ might be reached via other neighboring vertices $v'$ with lower path weight $d(v') + e_{v, v'}$.

We have put the new steps for the classification of \dijRect into \cref{alg:pathcnn} to better emphasize the difference to the standard Dijkstra's algorithm.
In an actual implementation, \dijRect only needs to be classified once for all neighbors of $u$.

\begin{figure}[!t]
  \centering
  \includegraphics[width=8.2cm]{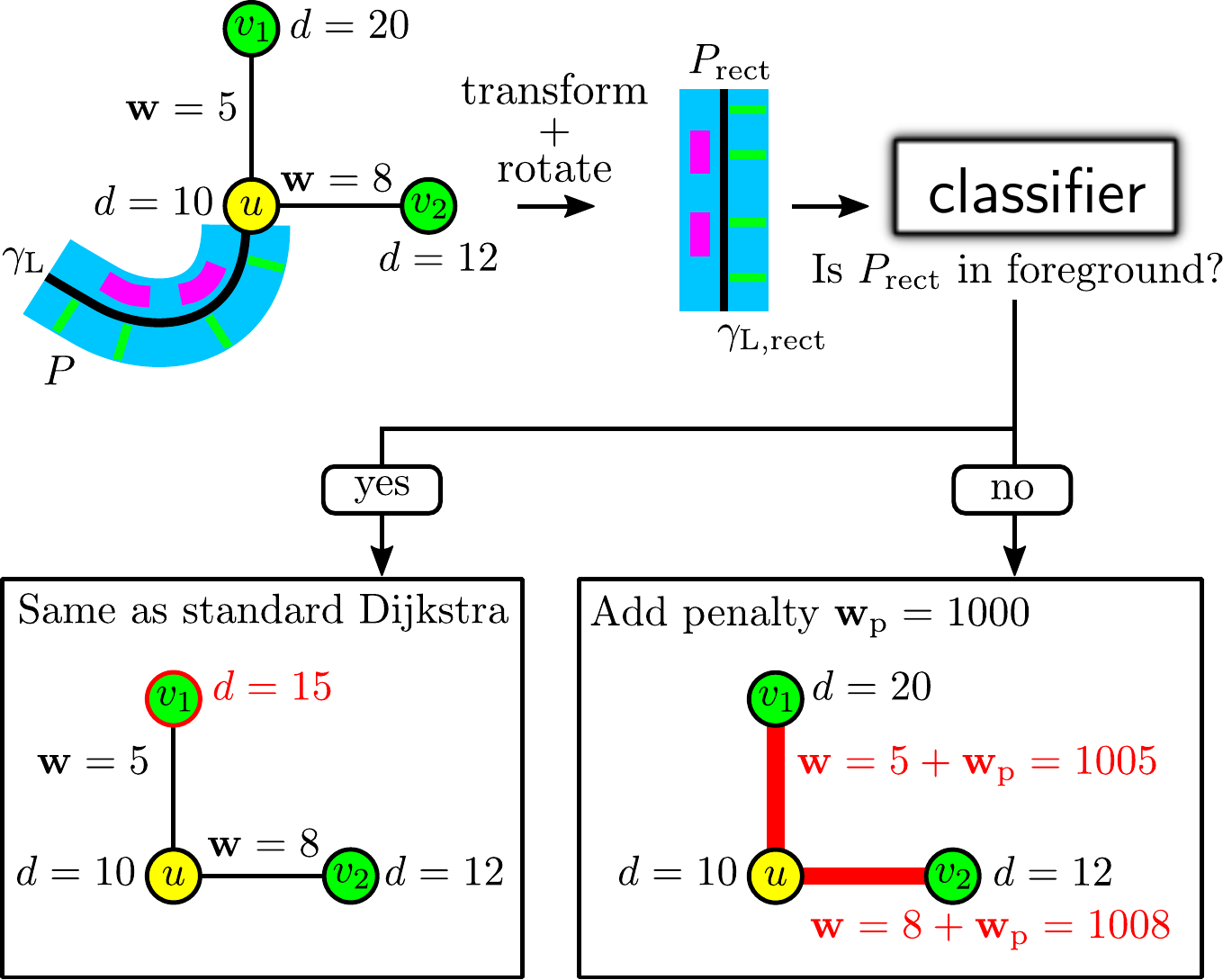}
  \vspace*{-5pt}
  \caption{
    An illustration of one step of the complete algorithm.
    Indices are omitted because they are obvious in the shown setting.
    Red color indicates changes compared to the previous step.
  }
  \label{pic:step}
\end{figure}

\paragraph{Illustration} One step of the main loop of the complete algorithm is demonstrated in \cref{pic:step}.
Suppose $u$ is the element in \dijQ with minimum path weight $\dijDist=10$, its neighbors in \dijQ have temporal path weights of $d(v_1)=20$ and $d(v_2)=12$, and the edge weights are $\dijW[e_{u,v_1}]=5$ and $\dijW[e_{u,v_2}]=8$.
After computing local path \dijPath (black line starting at $u$), the tubular patch \dijP can be extracted (blue stripe around \dijPath).
Magenta and green shapes inside \dijP represent its texture.
We then transform and rotate \dijP into a canonical orientation to obtain \dijRect, and \dijPath is accordingly transformed into a straight vertical line segment \dijPathRect in the middle of \dijRect.
The texture of \dijP undergoes the same transformation and rotation.
The CNN-classifier checks whether \dijRect is in foreground.
If this is the case, then $d(v_1)$ and $d(v_2)$ are updated as in the standard Dijkstra's algorithm:
The previous value of $d(v_1)$ is larger than $d(u) + \dijW[e_{u,v_1}] = 15$, so $d(v_1)$ is updated to $15$, and $\dijPrev(v_1) = u$.
$d(v_2)$ is not updated, since $d(u) + \dijW[e_{u,v_2}] = 18 > d(v_2)$.
In contrast, if \dijRect is \emph{not} in foreground, then \dijPath is a possible short cut.
Thus the weights $\dijW[e_{u,v_1}]$ and $\dijW[e_{u,v_2}]$ are both increased by a high penalty $\dijPenalty=1000$ to reduce the probability that $u$ becomes a vertex on the final minimal path.
In this case, neither $d(v_1)$ nor $d(v_2)$ changes, and \dijPrev is not updated.

\paragraph{F-maps} Our method not only provides centerlines of tubular structures, but also classifies pixels into foreground or background (\cref{func:classify} in \cref{alg:pathcnn}).
To avoid confusion with centerline segmentation, we use the term \emph{F-map} (foreground map) instead of \emph{segmentation} to refer to the image region classified as foreground.
F-maps have a remarkable property: Our CNN-classifiers are trained using only \emph{centerline} annotations, but the resulting F-maps provide \emph{binary segmentation masks} for the foreground.

\subsection{Training and Inference}
\label{sec:training_inference}

\begin{figure}[!htbp]
  \centering
  \begin{tabular}{cccc}
    \includegraphics[width=1.5cm]{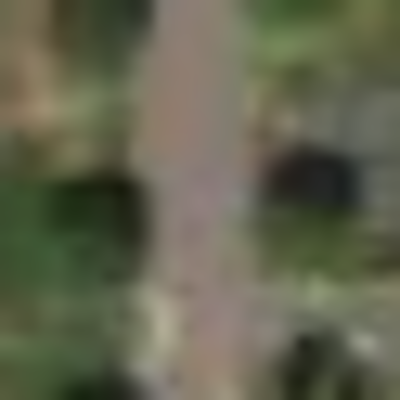}&
    \includegraphics[width=1.5cm]{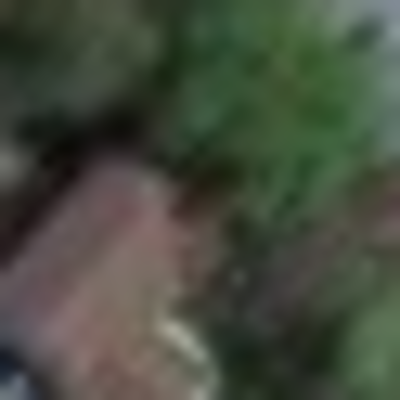}&
    \includegraphics[width=1.5cm]{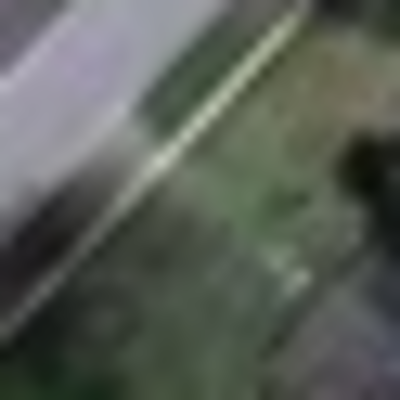}&
    \includegraphics[width=1.5cm]{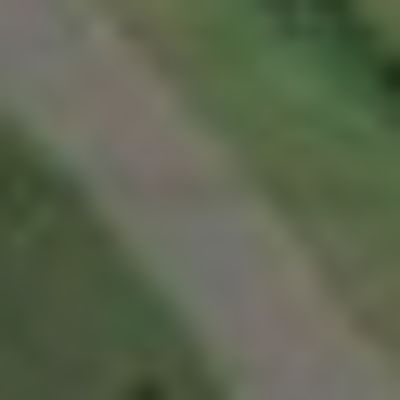}\\
    (a) Positive & (b) Negative & (c) Negative & (d) Negative
  \end{tabular}
\vspace*{-10pt}
  \caption{
    Samples for the training of our CNN.
    Positive samples must have a vertical road in the center of the rectified patch.
  }
  \label{pic:patches}
\end{figure}

To train the CNN, we use centerlines \emph{without} width information as annotations.
Rectified patches along these centerlines are used as positive samples, while negative samples are not rectified.
For example, in positive samples for road segmentation, there must be a vertical road in the center of the rectified patch (\cref{pic:patches}a).
In negative samples, either there is no road at all (\cref{pic:patches}b), or the road deviates strongly from the vertical position (\cref{pic:patches}c, d).
For data augmentation, rectified patches are only rotated in a very small range to compensate the possible numerical inaccuracies of centerlines.

The inference using CNN is embedded as the function \dijClassify in the minimal path computation in \cref{alg:pathcnn}.
The input images of the CNN are rectified patches along local paths, which are provided by the minimal path method.

\subsection{Interplay of Minimal Path Method and CNN}
The two main components of our method, i.e., the minimal path method and CNNs, benefit from each other to improve the overall centerline segmentation.

Using CNNs, local paths in the minimal path method can be better classified by exploiting strong classifiers to learn features automatically.
Also, given sufficient training samples, short cuts of Type 1 and 2 can be detected using a single classifier.
However, two difficulties remain:
CNNs cannot ensure topology of the centerline.
Also, during inference using our method, CNNs need rectified samples as input, instead of axis-aligned samples.
These samples must be first generated with other means.

The two difficulties of CNNs are addressed naturally using minimal path methods.
Line-topology of the centerlines is ensured as an inherent property of minimal path methods, and rectified patches can be obtained by computing local paths on-the-fly.
Rectification provides a strong geometric prior which significantly reduces the degree of freedom of input samples, i.e., the diversity of image patches caused by rotation and curvature is removed, since all paths become straight segments in the canonical orientation.
This has two advantages.
First, less data is needed to train the CNNs.
Second, it is easier for CNNs to classify paths, especially for complex shapes, such as paths with high curvature.

\section{Experimental Results}
We conducted experiments using four datasets of satellite images, including:
\dataepfl \cite{Sironi_PAMI_2015} (14 images of roads in suburban areas),
\dataroad (20 images of roads in rural areas),
\datariver (20 images of rivers in the wild), and
\datatoronto \cite{MnihThesis} (a road dataset with over 1000 high-resolution images).
Besides centerlines, \dataepfl contains also width information for roads, so we generated \emph{masks} of roads for this dataset.
The images in \dataroad and \datariver with centerline annotations were created by ourselves using data from GoogleEarth, since we were not aware of annotated datasets containing roads and rivers in such environments.

Our experiments are divided into two parts.
In the first part, we study the classification performance of the CNNs, and illustrate the impact of integrating CNNs into the minimal path framework.
In the second part, we apply our method to segment centerlines of tubular structures between user-specified start and end points, and compare our results with results of other minimal path methods and \abbrunet.

\subsection{Classification Performance of CNN}
\label{sec:cnn_performance}
In our method, the CNNs are used in two different ways compared to the usual application of CNNs.
First, our CNNs are applied only to \emph{rectified} patches, instead of arbitrary axis-aligned patches.
Second, our CNNs are embedded into minimal path methods.
In this section, we study the effects caused by these differences.

\begin{figure}[!htbp]
  \centering
  \begin{tabular}{cc}
    \includegraphics[width=3.8cm]{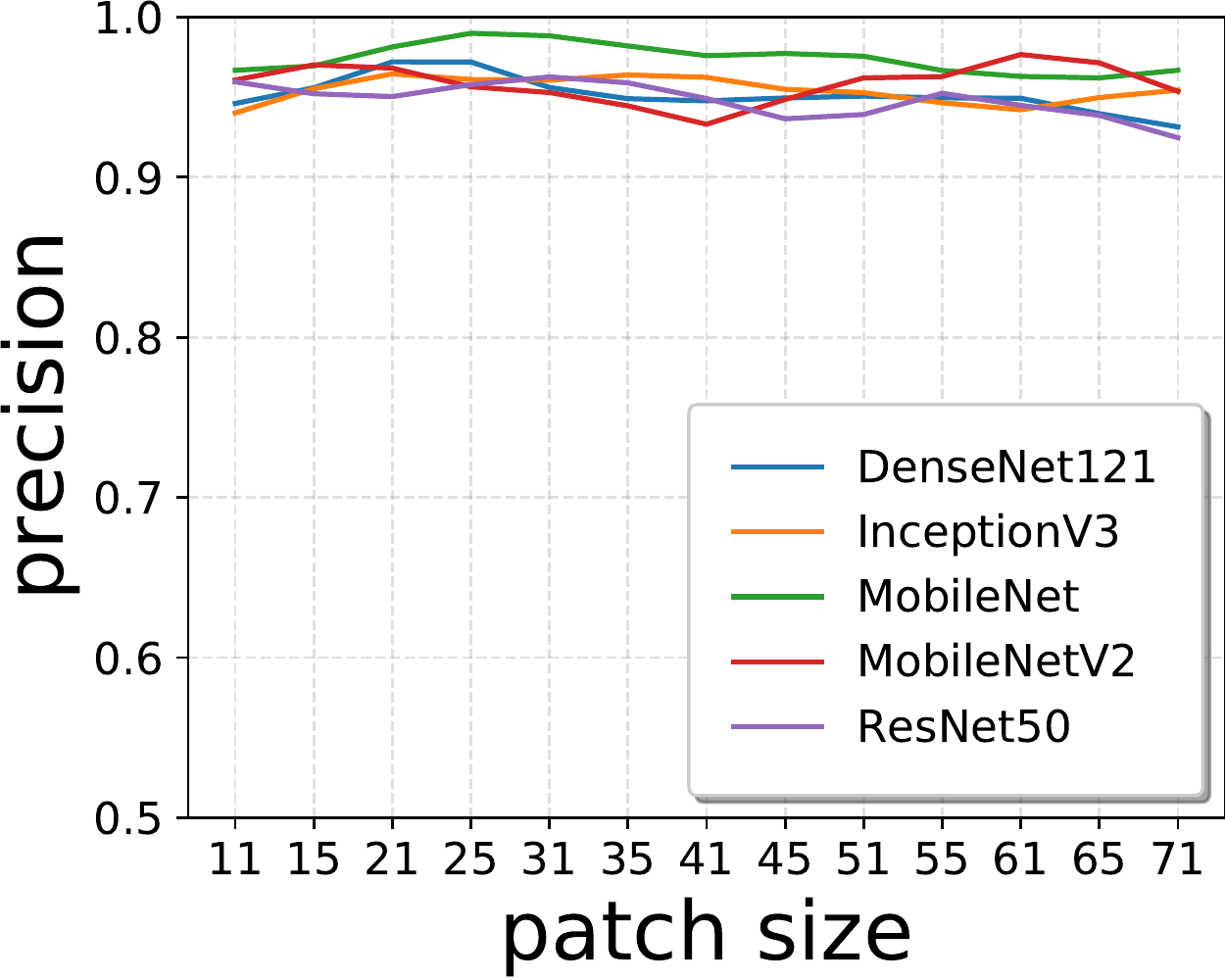}&
    \includegraphics[width=3.8cm]{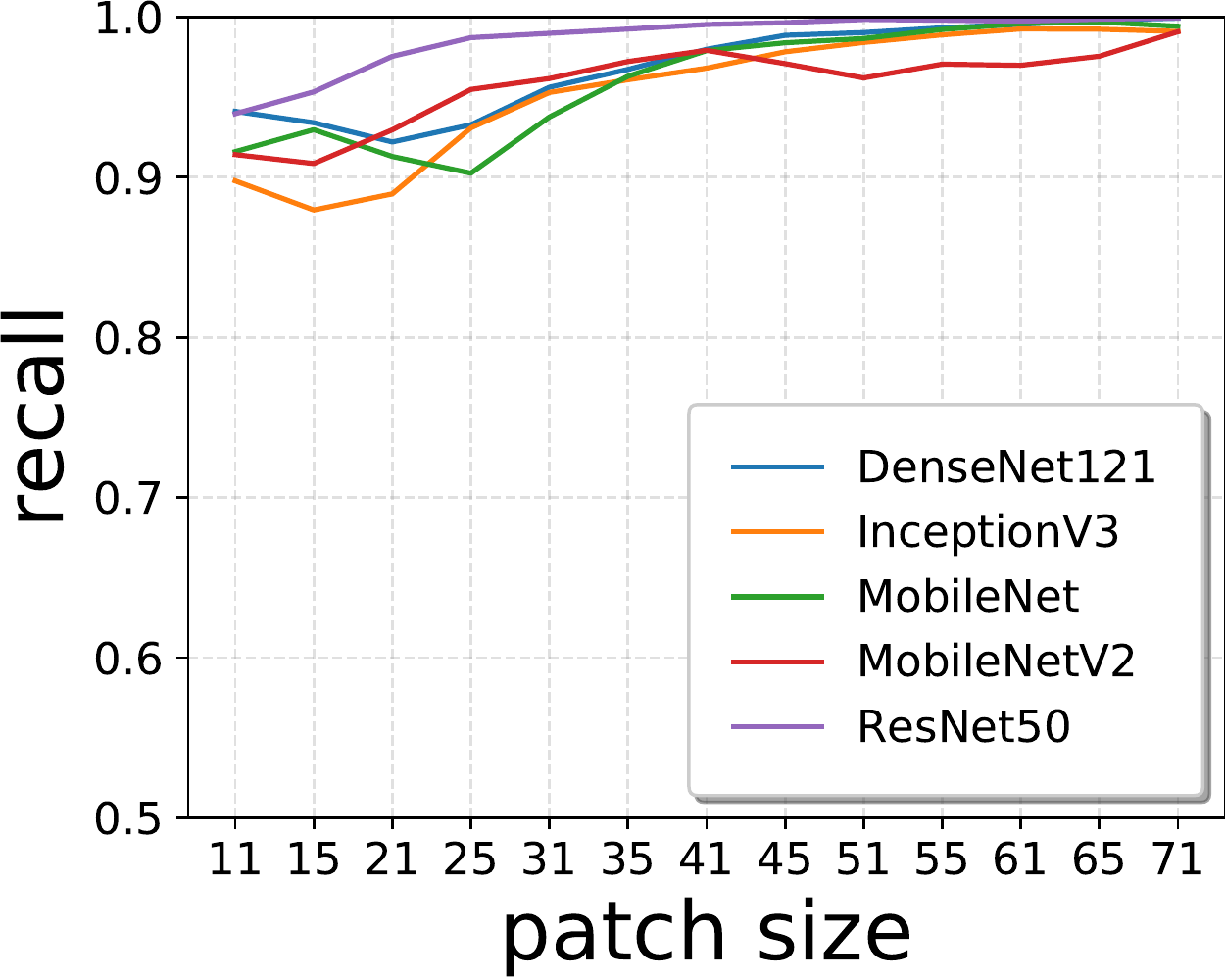}\\
  \end{tabular}
\vspace*{-10pt}
  \caption{
    Results of CNNs trained with rectified image patches.
  }
  \label{pic:statistics_patch_classification}
\end{figure}

The \dataepfl dataset was used for experiments in this section.
Samples from 5 images were used as training set, and samples from the other 9 images were used as test set.
For each experiment, we used roughly 1000 positive samples and 3000 negative samples for training.
The trained CNNs were tested on a test set of about 3000 rectified positive samples and 9000 negative samples.
The experiments were repeated for different sizes of the image patches, ranging from $11\times11$ to $71\times71$.
We tested different CNN architectures as well, including DenseNet121 \cite{Huang_CVPR_2017_DenseNet}, InceptionV3 \cite{Szegedy_CVPR_2016_InceptionV3}, MobileNet \cite{Howard_Arxiv_2017}, MobileNetV2 \cite{Sandler_CVPR_2018_MobileNetV2}, and ResNet50 \cite{He_CVPR_2016_ResNet}.
A comparison of them is shown in \cref{pic:statistics_patch_classification}.
It turns out that precision does \emph{not} always improve with increasing patch size, and the classification performance of different architectures does \emph{not} differ significantly.
Based on the trade-off between accuracy and speed, we selected MobileNet as our CNN, and consequently the maximum GPU memory consumption during testing is only about 2 GB.
We used patches of the size $31\times31$ for all experiments.

\begin{figure}[!tb]
  \hspace*{-0.4cm}
  \setlength{\tabcolsep}{3pt}
  \centering
  \begin{tabular}{ccc}
    \includegraphics[width=1.9cm]{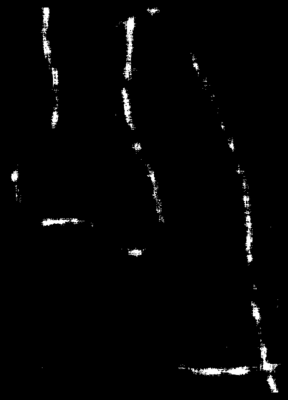}&
    \includegraphics[width=1.9cm]{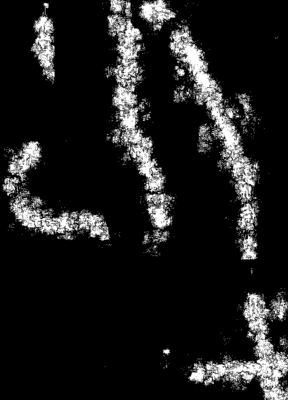}&
    \includegraphics[width=1.9cm]{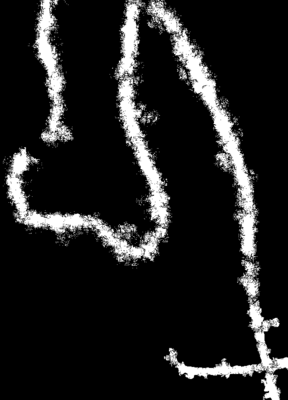}\\
    (a) Pixel-based CNN  & (b) No rectification & (c) Rectification
  \end{tabular}
\vspace*{-8pt}
  \caption{
    Regions classified as foreground.
    (a) $M_1$: Pixel-based classification.
    (b) $M_2$: Path classification, trained without rectification.
    (c) $M_3$: Path classification, trained using rectified samples.
  }
  \label{pic:epfl_shortcut_fmap}
\end{figure}

We use the image in \cref{pic:epfl_shortcut} from \dataepfl to compare three methods $M_1$, $M_2$, and $M_3$ for pixel classification.
$M_1$ and $M_2$ both employ the same classifier $C_a$ trained using axis-aligned patches, while $M_3$, i.e., our Path-CNN, uses a classifier $C_r$ trained using rectified patches.
$M_1$ applies $C_a$ directly to each pixel individually, while $M_2$ and $M_3$ both use our framework and embed the classifier into the Dijkstra's algorithm.
With our sparse training data, $M_1$ results in a large number of gaps, and the tubular structures are difficult to recognize (\cref{pic:epfl_shortcut_fmap}a).
In contrast, by embedding the same CNN $C_a$ into our minimal path method, the F-map of $M_2$ shows much more geometric details of the roads (\cref{pic:epfl_shortcut_fmap}b).
Trained with rectified patches, the F-map of $M_3$ contains even less noise and discontinuities in the tubular structures (\cref{pic:epfl_shortcut_fmap}c).
The maps in \cref{pic:epfl_shortcut_fmap}a, b, and c lead to the blue, green, and red paths in \cref{pic:epfl_shortcut}b, respectively.

\subsection{Centerline Segmentation}
\label{sec:result_centerline}

\begin{table}[t]
  \begin{center}
    \begin{tabular}{c|c|c|c|c}
      \hline
       & \textbf{\dataepfl} & \textbf{\dataroad} & \textbf{\datariver} & \textbf{\datatoronto} \\\hline\hline
      Train.~images&5&6&6&3 (12)\\\hline
      Test images&9&14&14&49\\\hline
      Test paths&130&64&14&751\\\hline
    \end{tabular}
  \end{center}
  \vspace*{-10pt}
  \caption{Overview of data used in experiments.
    }
  \label{table:training_data}
\end{table}
\paragraph{Data} The overview of data for our experiments are shown in \cref{table:training_data}.
For each test path, we set start and end points manually.
Our training sets are quite small.
For example, for \datatoronto we used patches from 12 images, which in total amount to the size of only 3 full images.
In contrast, recent approaches \cite{Bastani_CVPR_2018, Mosinska_PAMI_2020} both used 25 images for training and 15 for testing.
Our method is trained only using centerlines without width information.
\paragraph{Baselines} We compared our method with five other approaches, including \abbrplain, \dcnn, \abbrprog, \pcnn, and \abbrunet \cite{Ronneberger2015}.
\abbrplain and \dcnn are the standard Dijkstra's algorithm \cite{Dijkstra59:NM} using different tubularity measures: \abbrplain uses the measure \cite{Frangi1998}, while \dcnn uses MobileNet features (i.e., pixel-wise classification results of a MobileNet like the $M_1$ method in \cref{sec:cnn_performance}).
Similarly, \abbrprog and \pcnn are the progressive minimal path method \cite{Liao_PAMI_2018_Final} using the measure \cite{Frangi1998} and MobileNet features, respectively.
Like \abbrplain and \abbrprog, our method also uses the tubularity measure \cite{Frangi1998} to compute the initial edge weight \dijWinit.
In \abbrplain and \dcnn, the edge weight remains constant, while \abbrprog and \pcnn adapt the edge weight by classifying local paths.
But unlike our approach, there is no cropping or rectification in \abbrprog or \pcnn.
Instead, only mean values of the tubularity measure \cite{Frangi1998} or MobileNet features on the local paths are used for classification.
Similar to the definition of F-map for our method, we define the F-map of \abbrprog and \pcnn as regions classified as foreground.
\abbrplain and \dcnn have no classification step, therefore we define the F-map of \abbrplain and \dcnn as regions for which the distance to the start point will not change anymore (i.e., the region \dijS in \cref{alg:dijkstra}) when the end point is reached.
\paragraph{Error analysis} To measure the errors of the results quantitatively, we used the mean distance between segmented centerlines and the corresponding ground truth, defined as:
\begin{equation}
\errormeasure = \frac{1}{N}\sum_{\mypath}\sum _{\position_i\in \mypath}\singleerrormeasure(\position_i) = \frac{1}{N}\sum_{\mypath}\sum _{\position_i\in \mypath} |\position_i - \refpoint{i}|,
\label{eqn:error_measure}
\end{equation}
where $N$ is the total number of points on all segmented centerlines \mypath, $\position_i$ are the coordinates of points on \mypath, and $\refpoint{i}$ are the closest points to $\position_i$ on the ground truth.
The results are summarized in \cref{table:mean_dist}.
For all datasets, our method achieved the lowest mean distances.
The errors of \abbrplain and \abbrprog are higher than our results by a factor of at least 2.
\dcnn and \pcnn achieve lower errors, which are still significantly higher than those of our results.
\begin{table*}[t]
  \begin{center}
    \begin{tabular}{c|c|c|c|c|c}
      \hline
      \textbf{Dataset} & \textbf{\abbrplain}     &\textbf{\abbrprog}       & \textbf{\dcnn}         & \textbf{\pcnn}        & \textbf{Our method}\\\hline\hline
      \dataepfl        & 4.74 \errorratio{2.82}  & 3.36 \errorratio{2.00}  & 2.82 \errorratio{1.68} & 2.74 \errorratio{1.63} & \textbf{1.68}\\\hline
      \dataroad        & 6.69 \errorratio{2.25}  & 5.98 \errorratio{2.01}  & 3.91 \errorratio{1.32} & 3.68 \errorratio{1.24} & \textbf{2.97}\\\hline
      \datariver       & 19.01 \errorratio{5.03} & 18.43 \errorratio{4.88} & 7.16 \errorratio{1.89} & 6.31 \errorratio{1.67} & \textbf{3.78}\\\hline
      \datatoronto     & 21.89 \errorratio{6.10} & 18.83 \errorratio{5.25} & 6.17 \errorratio{1.72} & 5.25 \errorratio{1.46} & \textbf{3.59}\\\hline
    \end{tabular}
  \end{center}
\vspace*{-10pt}
  \caption{Errors measured with \cref{eqn:error_measure} in pixels.
    Numbers in braces indicate ratios between results of the other methods and ours.}
  \label{table:mean_dist}
\end{table*}
\begin{figure*}[!htbp]
  \centering
  \hspace*{-10pt}
  \begin{tabular}{cccc}
    \includegraphics[width=4cm]{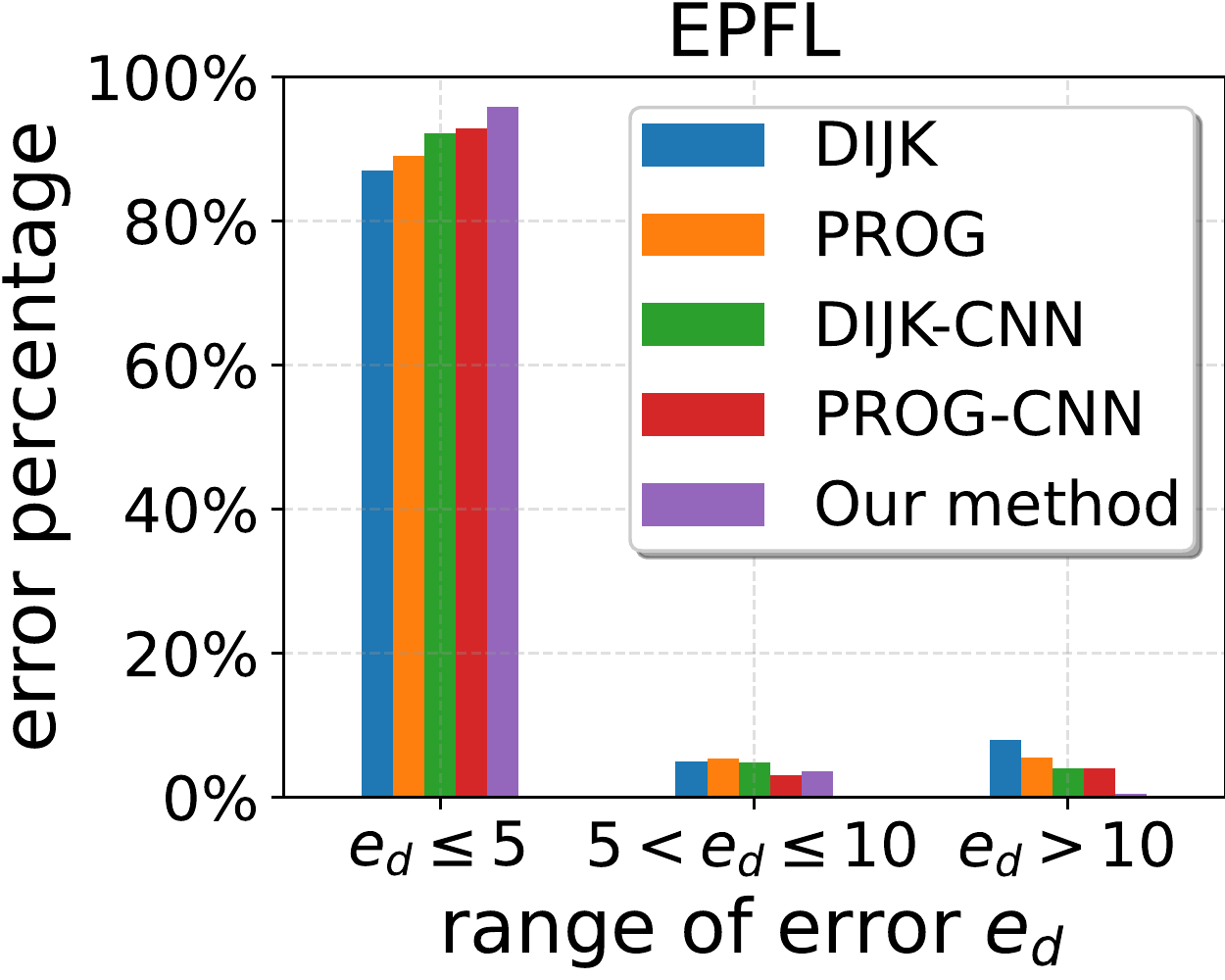}&
    \includegraphics[width=4cm]{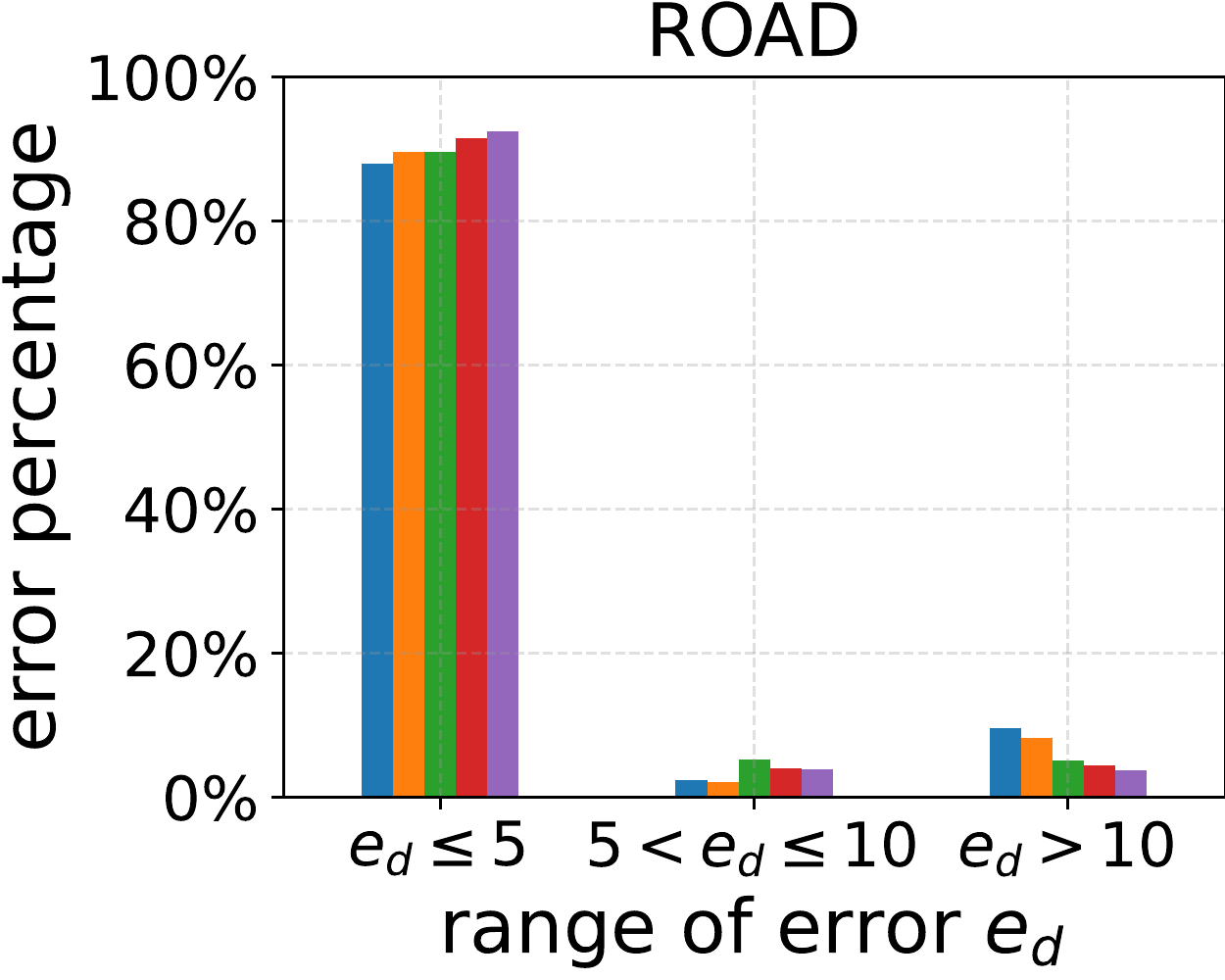}&
    \includegraphics[width=4cm]{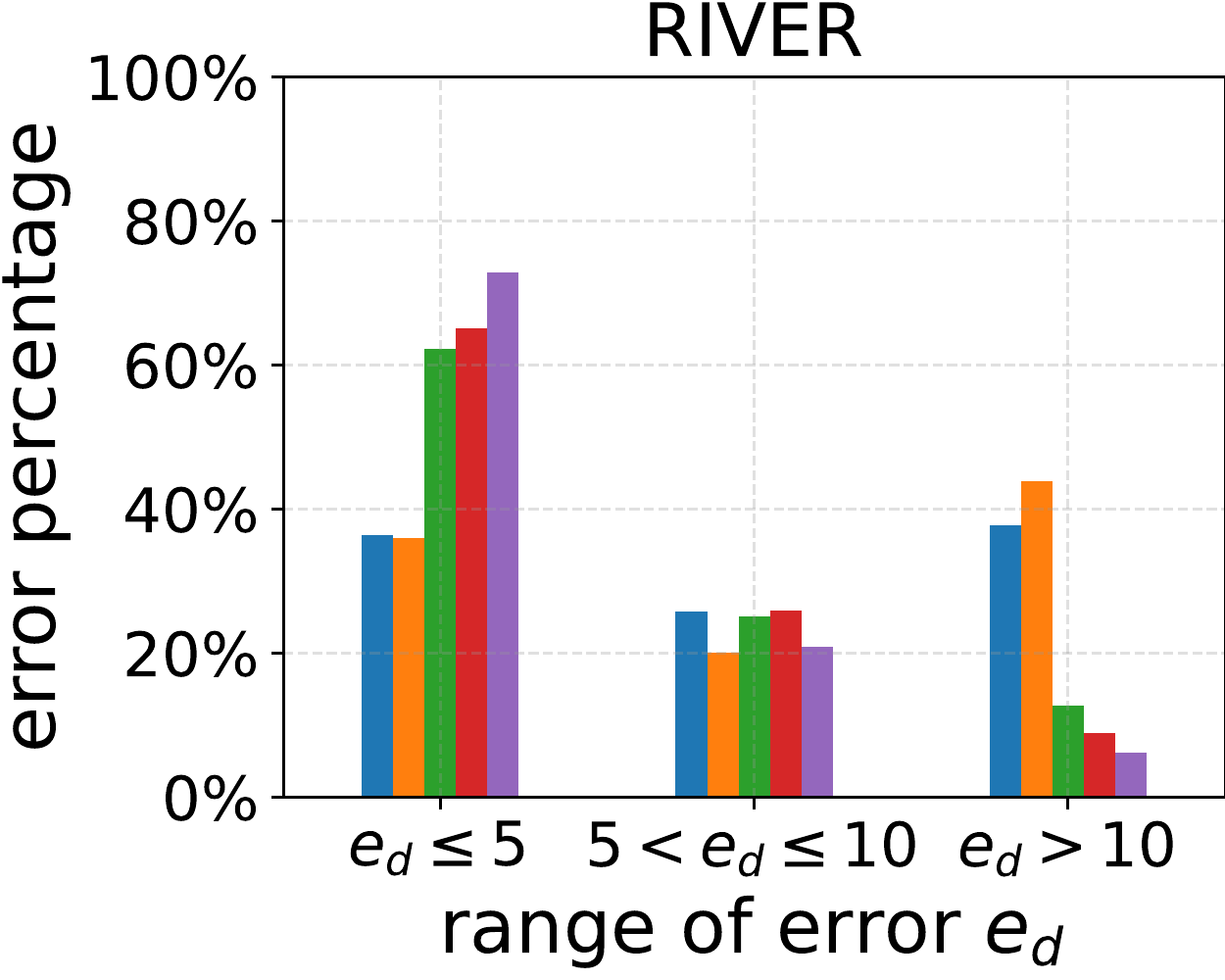}&
    \includegraphics[width=4cm]{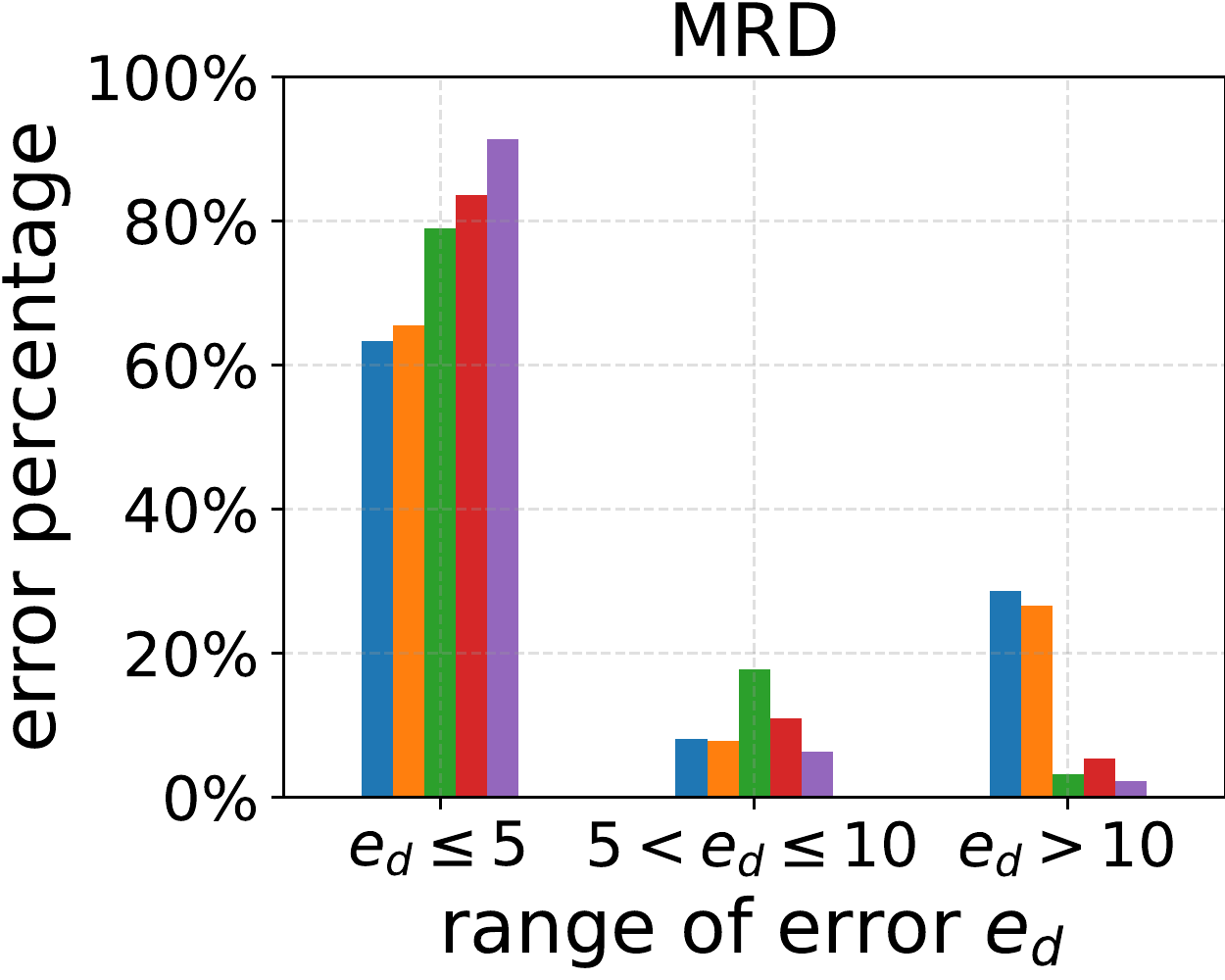}\\
  \end{tabular}
\vspace*{-10pt}
  \caption{
    Distribution of errors for centerline segmentation using four datasets.
  }
  \label{pic:statistics_error_distribution}
\end{figure*}

In \cref{pic:statistics_error_distribution}, the errors $\singleerrormeasure$ for points on the segmented centerlines are divided into three intervals, i.e., errors smaller than 5 pixels, between 5 and 10 pixels, and larger than 10 pixels, respectively.
The lowest error bound is set to 5 pixels, because this is roughly half the width of the tubular structures in our images.
A centerline within this range is considered close enough to the ground truth, since we do not use width information of the tubular structures to train our algorithm.
Intuitively, higher percentage of large errors indicate more short cuts.
Using our method, most points on the centerlines are within smaller error ranges.
For example, among the points on the centerlines segmented by our method in \dataepfl, $95.78\%$ have errors under 5 pixels, and only $0.52\%$ have errors larger than 10 pixels.
In contrast, \abbrplain, \abbrprog, \dcnn and \pcnn have $4\%$ to $8\%$ errors larger than 10 pixels.
For \datariver, $72.89\%$ of the errors of our method are under 5 pixels.
The percentage of errors over 10 pixels is $6.20\%$ for our method, but $37.71\%$, $43.88\%$, $12.71\%$ and $8.92\%$ for \abbrplain, \abbrprog, \dcnn and \pcnn, respectively.
Also for \datatoronto, our method has $91.38\%$ errors less than 5 pixels.

\begin{table}[t]
  \begin{center}
    \begin{tabular}{c|c|c|c}
      \hline
      & \textbf{\abbrunet } & \textbf{\abbrunet } & \textbf{Our method}\\\hline\hline
      Annotation type & centerline&mask&centerline \\\hline
      Dice coefficient& 0.1163&0.2953 & \textbf{0.6108}\\\hline
    \end{tabular}
  \end{center}
  \vspace*{-10pt}
  \caption{Dice coefficients of \abbrunet and our method for \dataepfl
  }
  \label{table:conv_vs_unet}
\end{table}

\begin{figure}[!htbp]
  \centering
  \begin{tabular}{ccc}
    \includegraphics[width=1.9cm]{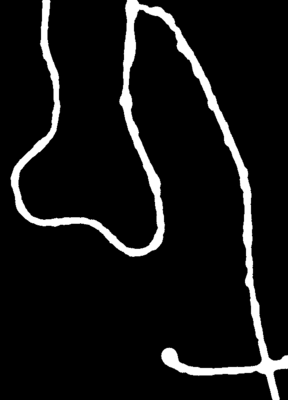}&
    \includegraphics[width=1.9cm]{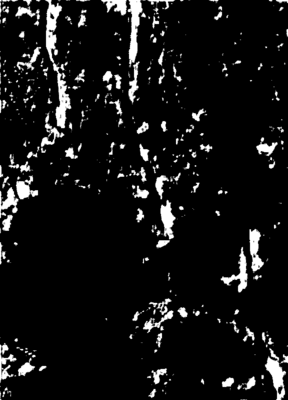}&
    \includegraphics[width=1.9cm]{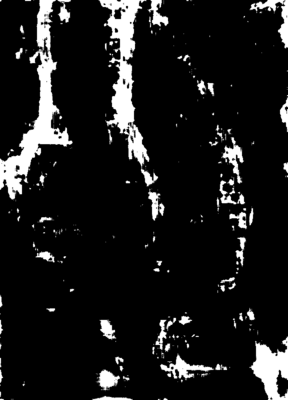}
  \end{tabular}
  \vspace*{-10pt}
  \caption{
    Segmentation results of \abbrunet for the image in \cref{pic:epfl_shortcut}a.
    Left: Ground truth.
    Middle: \abbrunet trained with only centerlines (Dice $= 0.1511$).
    Right: \abbrunet trained with masks (Dice $= 0.3521$).
    Our result is shown in \cref{pic:epfl_shortcut_fmap}c (Dice $=0.6658$).
  }
  \label{pic:conv_vs_unet}
\end{figure}

As masks for roads are available for \dataepfl, we computed Dice coefficients for \abbrunet trained with only centerlines, \abbrunet trained with masks, and our method.
Results for the 9 test images are shown in \cref{table:conv_vs_unet}.
Although our method was trained using only centerlines, it achieved the highest Dice coefficient.
An example is shown in \cref{pic:conv_vs_unet}.

\paragraph{Qualitative results}
Below we show more results exemplarily.
In the figures, the start points and end points are shown as magenta boxes and cyan circles, respectively.

\begin{figure}[!tb]
  \centering
  \begin{tabular}{cc}
    \includegraphics[height=2.5cm]{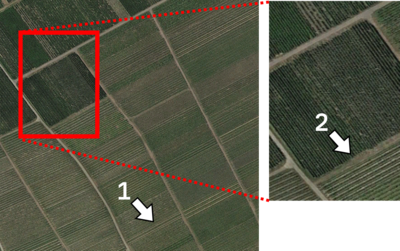}&
    \includegraphics[height=2.5cm]{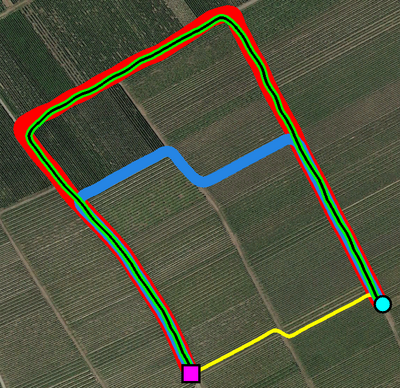}\\
    (a) Input image &  (b) Results
  \end{tabular}
\vspace*{-9pt}
  \caption{Results for one image from \dataroad.
    (a) The white arrows show two difficult cases.
    (b) Results of \abbrplain, \abbrprog, \dcnn, \pcnn and our method are shown in yellow, blue, green, black, and red colors, respectively.
  }
  \label{pic:road1_08}
\end{figure}

\begin{figure*}
  \centering
  \begin{tabular}{ccccccc}
    \includegraphics[width=2.1cm]{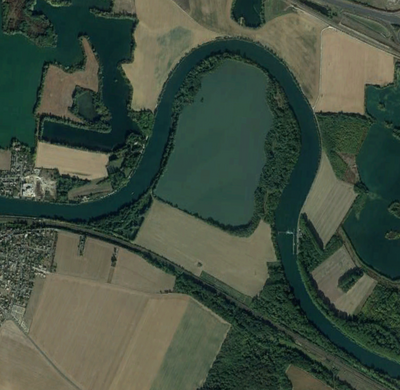}&
\includegraphics[width=2.1cm]{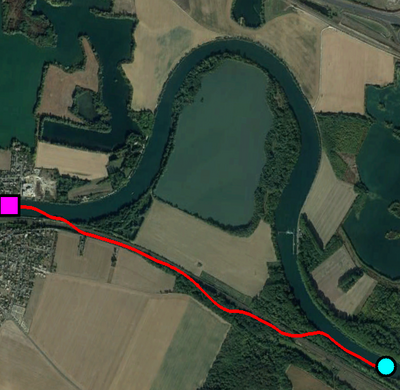}&
\includegraphics[width=2.1cm]{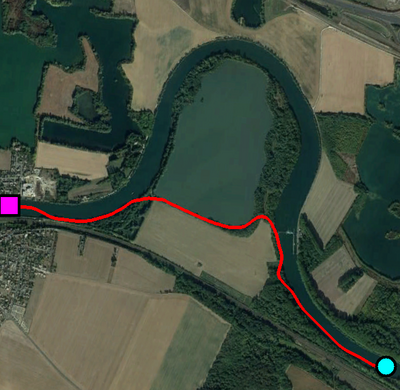}&
\includegraphics[width=2.1cm]{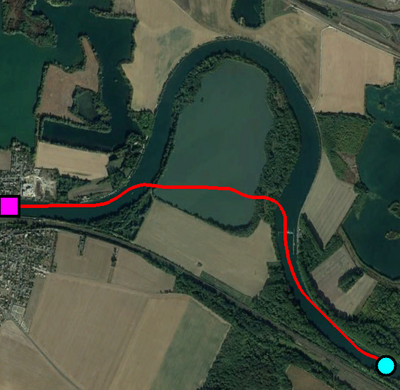}&
\includegraphics[width=2.1cm]{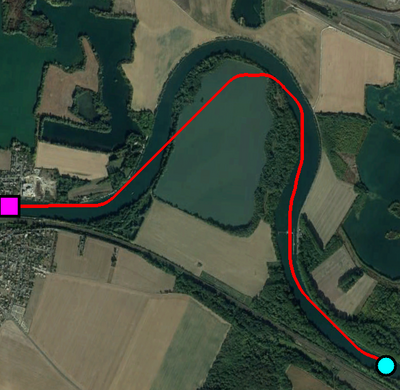}&
\includegraphics[width=2.1cm]{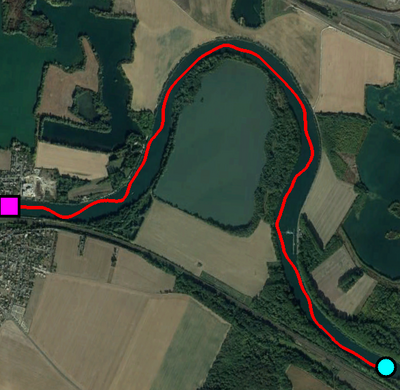}&
\includegraphics[width=2.1cm]{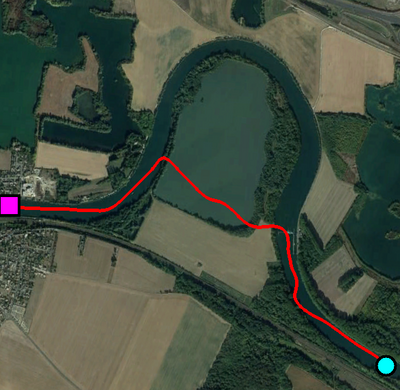}\\
\includegraphics[width=2.1cm]{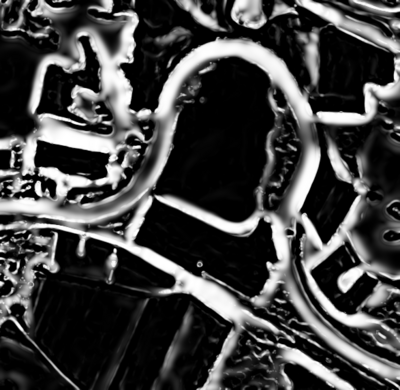}&
\includegraphics[width=2.1cm]{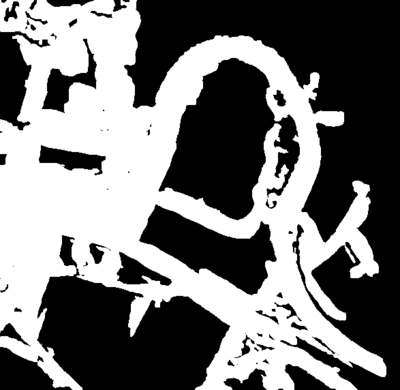}&
\includegraphics[width=2.1cm]{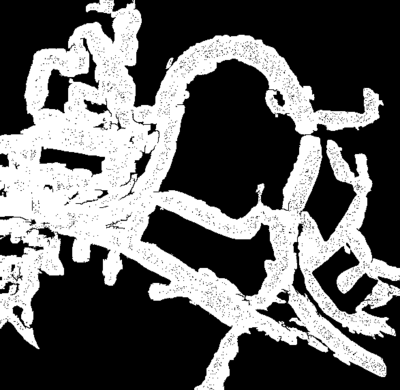}&
\includegraphics[width=2.1cm]{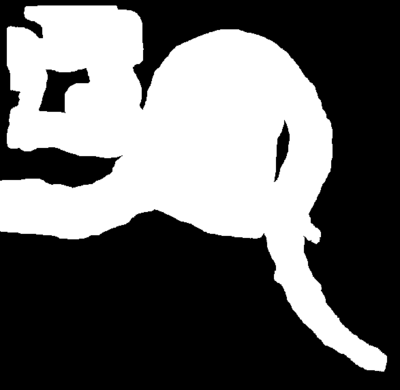}&
\includegraphics[width=2.1cm]{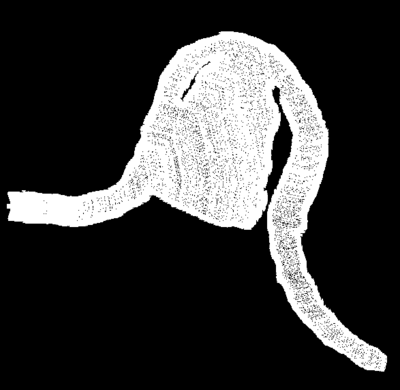}&
\includegraphics[width=2.1cm]{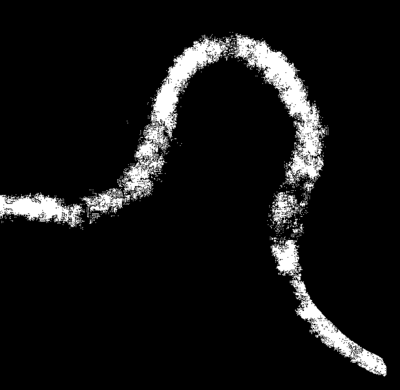}&
\includegraphics[width=2.1cm]{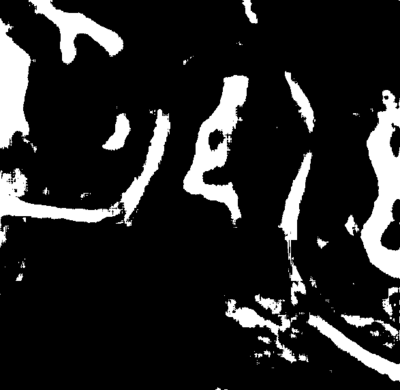}\\
(a) \small{Input} & (b) \small{\abbrplain} & (c) \small{\abbrprog} & (d) \small{\dcnn} & (e) \small{\pcnn} & (f) \small{\textbf{Our method}} & (g) \small{U-Net}
  \end{tabular}
  \vspace*{-8pt}
  \caption{Results for one image from \datariver.
    (a) Input image and tubularity map.
    (b) - (f) Results of \abbrplain, \abbrprog, \dcnn, \pcnn, our method (upper row) and the corresponding F-maps (lower row).
    (g) Results using U-Net.
  }
  \label{pic:river1_08}
\end{figure*}

The images in \dataroad and \datariver are challenging since there are a lot of tubular structures in the background.
Although these structures have slightly different appearance than the actual roads or rivers, it is difficult to use tubularity feature alone to capture the differences.
For an image from \dataroad, two difficult cases are highlighted in \cref{pic:road1_08} (white arrows).
In case 1, \abbrplain found a short cut (Type 1), because it is much shorter than the correct path.
\abbrprog was able to detect this short cut and avoid it.
But in case 2, where the tubularity measure in the background is also high, \abbrprog found a short cut (Type 2).
In contrast, \dcnn, \pcnn and our approach handled both cases correctly and found the correct centerline.
\cref{pic:river1_08} shows a more challenging example from \datariver, and only our method successfully segmented the river.
The F-maps show that while our method correctly classified most image regions and mostly remained in the foreground, all other approaches explored large portions of background, and resulted in short cuts.
We also computed the path by applying \abbrplain to the mask obtained using \abbrunet instead of the tubularity map.
However, with the small number of samples in our training set (only 9 rivers in the 6 training images), \abbrunet could not be trained well enough to produce a good mask for foreground.
Consequently, the correct path was not found (\cref{pic:river1_08}g).

\begin{figure}[!tb]
  \hspace*{-0.4cm}
  \setlength{\tabcolsep}{3pt}
  \centering
  \begin{tabular}{ccc}
    \includegraphics[height=2.6cm]{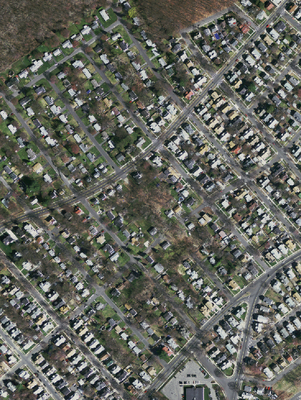}&
    \includegraphics[height=2.6cm]{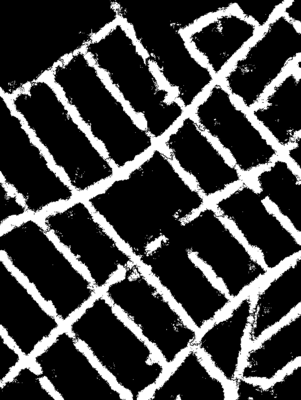}&
    \includegraphics[height=2.6cm]{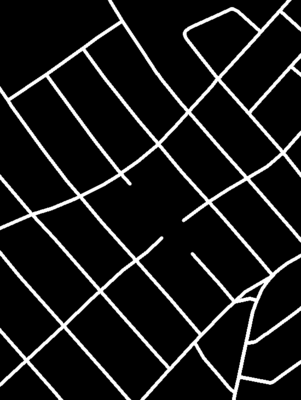}\\
    (a) Input image & (b) F-map & (c) Ground truth
    \end{tabular}
  \vspace*{-8pt}
  \caption{
    F-map (mask) for a section of an image from \datatoronto.
    The ground truth contains only centerlines, not masks.
  }
  \label{pic:toronto_section}
\end{figure}

\begin{figure}[!tb]
  \hspace*{-0.4cm}
  \setlength{\tabcolsep}{3pt}
  \centering
  \begin{tabular}{ccc}
    \includegraphics[width=2.6cm]{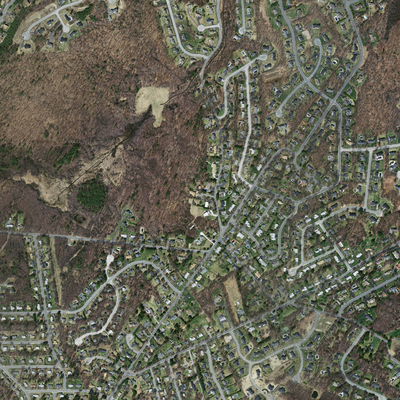}&
    \includegraphics[width=2.6cm]{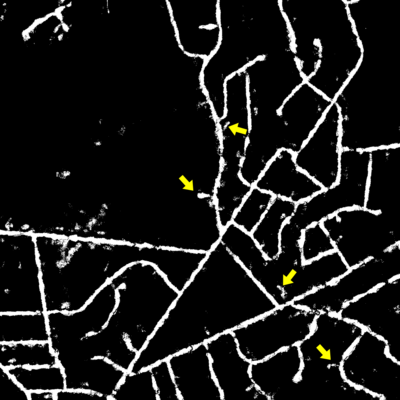}&
    \includegraphics[width=2.6cm]{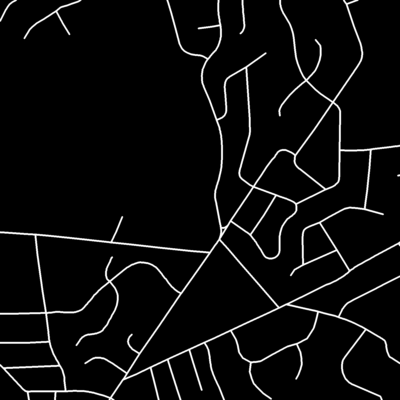}\\
    (a) Input image  & (b) F-map & (c) Ground truth
  \end{tabular}
\vspace*{-8pt}
  \caption{
    F-map for a full image from \datatoronto obtained using one start point for the entire image.
    Some roads, which are missing in the ground truth, were detected using our method (yellow arrows).
  }
  \label{pic:toronto_full}
\end{figure}

Images in \datatoronto cover large areas with complex road networks.
The roads may appear very differently even within the same image, and they are often occluded by trees or their shadows.
Buildings along the roads frequently show certain regular patterns, causing high tubularity outside the roads.
There are also tubular structures which are not roads, such as rivers, or narrow spaces between buildings.
\cref{pic:toronto_section} shows a section of an image of an urban area.
There, our method dealt well with the challenges above, yielding a quite precise F-map (\cref{pic:toronto_section}b).
\emph{Without re-training}, the same model can be used for roads in other environments.
For example, the F-map of a full image of a suburban area is shown in \cref{pic:toronto_full}.
Our methods even detected several small roads which are missing in the ground truth.
To obtain the F-map for all roads in the image, the user only needs to set \emph{one single start point} for the entire image, i.e., no end points or further start points are needed, even if the roads are not connected with each other.
F-maps are \emph{not} sensitive to the location of the start point, so similar F-maps can be obtained using different start points.

\section{Conclusion}
We introduced Path-CNN, a new method for topology-aware centerline segmentation for tubular structures. In our method, a CNN is embedded as an integral component into the progressive minimal path method.
The CNN enhances hand-tuned image features to better control the minimal path computation, while the progressive minimal path method provides strong geometric priors to improve the performance of the CNN, and ensures the line-topology of the segmented centerlines.
Path-CNN employs path-based classification to avoid different types of short cuts, and consequently centerlines can be better segmented, especially for tubular structures with complex shapes in challenging environments.
In addition to centerline segmentation, a binary mask (F-map) is also obtained.
Our method only needs sparse and simple annotations for training, and it has lower hardware requirements than many other methods.
Its effectiveness is demonstrated using experiments with four datasets of satellite images and comparison with five other methods.
Future work includes extension of our method for medical images, especially 3D medical images.

{\small
  \bibliographystyle{ieee_fullname}
  \bibliography{literature,egbib}
}

\end{document}